%% file: main.tex
\documentclass{article}
\usepackage{amssymb}

\usepackage[final]{corl_2025} 

\usepackage{times}
\usepackage[numbers]{natbib}
\usepackage{multicol}
\usepackage{wrapfig}
\usepackage{booktabs}
\usepackage[table,xcdraw]{xcolor}
\usepackage{colortbl}
\usepackage{graphicx}
\usepackage{amsmath}
\usepackage{amsfonts}
\usepackage[ruled,vlined,linesnumbered]{algorithm2e}
\usepackage{cuted} 
\usepackage{caption}
\usepackage{lipsum}  
\usepackage{overpic}
\usepackage{algorithmicx}
\usepackage{tabularx, makecell} 

\title{Learning from 10 Demos:\\Generalisable and Sample-Efficient Policy Learning with Oriented Affordance Frames}

\author{
    Krishan Rana$^{\dagger1}$, Jad Abou-Chakra$^{1}$, Sourav Garg$^{2}$, Robert Lee,\\
    \textbf{Ian Reid}$^2$, \textbf{Niko S\"underhauf}$^1$ \\
    $^1$QUT Centre for Robotics, Queensland University of Technology \\
    $^2$University of Adelaide \\    $^{\dagger}$\texttt{ranak@qut.edu.au}
}

\begin{document}
\maketitle


\begin{abstract}
Imitation learning has unlocked the potential for robots to exhibit highly dexterous behaviours. However, it still struggles with long-horizon, multi-object tasks due to poor sample efficiency and limited generalisation. Existing methods require a substantial number of demonstrations to cover possible task variations, making them costly and often impractical for real-world deployment. 
We address this challenge by introducing \emph{oriented affordance frames}, a structured representation for state and action spaces that improves spatial and intra-category generalisation and enables policies to be learned efficiently from only 10 demonstrations. More importantly we show how this abstraction allows for compositional generalisation of independently trained sub-policies to solve long-horizon, multi-object tasks. To seamlessly transition between sub-policies, we introduce the notion of self-progress prediction, which we directly derive from the duration of the training demonstrations. We validate our method across three real-world tasks, each requiring multi-step, multi-object interactions. Despite the small dataset, our policies generalise robustly to unseen object appearances, geometries, and spatial arrangements, achieving high success rates without reliance on exhaustive training data. Video demonstration can be found on our project page: \href{https://affordance-policy.github.io/}{https://affordance-policy.github.io/}.

\end{abstract}

\keywords{behaviour cloning, imitation learning, generalisation, affordances} 

\input{mathstuff}

\input{introduction.tex}

\input{related_work}

\input{method}
\input{evaluation.tex}

\input{discussion}

\bibliography{bibliography}

\input{appendix}

\end{document}

%% file: mathstuff.tex
\newcommand{\vect}[1]{\mathbf{ #1}}
\newcommand{\vectg}[1]{{\boldsymbol{ #1}}}
\newcommand{\ggo}{\ensuremath{\mathrm{g^2o}} }
\newcommand{\R}{\mathbb{R}}
\newcommand{\N}{\mathbb{N}}
\newcommand{\Z}{\mathbb{Z}}
\renewcommand{\P}{\mathbb{P}}
\newcommand{\tran}{^\top}
\newcommand{\T}{^\mathsf{T}}
\newcommand{\iT}{^{-\mathsf{T}}}
\newcommand{\inv}{^{-1}}
\newcommand{\func}[2]{\mathtt{#1}\left\{#2\right\}}
\newcommand{\sig}{\operatorname{sig}}
\newcommand{\diag}{\operatorname{diag}}
\newcommand{\argmin}{\operatornamewithlimits{argmin}}
\newcommand{\argmax}{\operatornamewithlimits{argmax}}
\newcommand{\RMSE}{\operatorname{RMSE}}
\newcommand{\RMSEpos}{\operatorname{RMSE}_\text{pos}}
\newcommand{\RMSEori}{\operatorname{RMSE}_\text{ori}}
\newcommand{\RPE}{\operatorname{RPE}}
\newcommand{\RPEpos}{\operatorname{RPE}_\text{pos}}
\newcommand{\RPEori}{\operatorname{RPE}_\text{ori}}
\newcommand{\rpe}{\varepsilon_{\vdelta}}
\newcommand{\achiError}{\bar{e}_{\chi^2}}
\newcommand{\chiError}{e_{\chi^2}}
\newcommand{\normal}[2]{\mathcal{N}\left(#1, #2\right)}
\newcommand{\uniform}[2]{\mathcal{U}\left(#1, #2\right)}
\newcommand{\pfrac}[2]{\frac{\partial #1}{\partial #2}}  
\newcommand{\fracpd}[2]{\frac{\partial #1}{\partial #2}} 
\newcommand{\fracppd}[2]{\frac{\partial^2 #1}{\partial #2^2}}  
\newcommand{\dd}{\mathrm{d}}  
\newcommand{\smd}[2]{\left\| #1 \right\|^2_{#2}}
\newcommand{\E}[1]{\text{\normalfont{E}}\left[ #1 \right]}     
\newcommand{\Cov}[1]{\text{\normalfont{Cov}}\left[ #1 \right]} 
\newcommand{\Var}[1]{\text{\normalfont{Var}}\left[ #1 \right]} 
\newcommand{\Tr}[1]{\text{\normalfont{tr}}\left( #1 \right)}   
\def\sgn{\mathop{\mathrm sgn}}    
\newcommand{\twovector}[2]{\begin{pmatrix} #1 \\ #2 \end{pmatrix}} 
\newcommand{\smalltwovector}[2]{\left(\begin{smallmatrix} #1 \\ #2 \end{smallmatrix}\right)} 
\newcommand{\threevector}[3]{\begin{pmatrix} #1 \\ #2 \\ #3 \end{pmatrix}} 
\newcommand{\fourvector}[4]{\begin{pmatrix} #1 \\ #2 \\ #3 \\ #4 \end{pmatrix}}  
\newcommand{\smallthreevector}[3]{\left(\begin{smallmatrix} #1 \\ #2 \\ #3 \end{smallmatrix}\right)} 
\newcommand{\fourmatrix}[4]{\begin{pmatrix} #1 & #2 \\ #3 & #4 \end{pmatrix}} 
\newcommand{\vA}{\vect{A}}
\newcommand{\vB}{\vect{B}}
\newcommand{\vC}{\vect{C}}
\newcommand{\vD}{\vect{D}}
\newcommand{\vE}{\vect{E}}
\newcommand{\vF}{\vect{F}}
\newcommand{\vG}{\vect{G}}
\newcommand{\vH}{\vect{H}}
\newcommand{\vI}{\vect{I}}
\newcommand{\vJ}{\vect{J}}
\newcommand{\vK}{\vect{K}}
\newcommand{\vL}{\vect{L}}
\newcommand{\vM}{\vect{M}}
\newcommand{\vN}{\vect{N}}
\newcommand{\vO}{\vect{O}}
\newcommand{\vP}{\vect{P}}
\newcommand{\vQ}{\vect{Q}}
\newcommand{\vR}{\vect{R}}
\newcommand{\vS}{\vect{S}}
\newcommand{\vT}{\vect{T}}
\newcommand{\vU}{\vect{U}}
\newcommand{\vV}{\vect{V}}
\newcommand{\vW}{\vect{W}}
\newcommand{\vX}{\vect{X}}
\newcommand{\vY}{\vect{Y}}
\newcommand{\vZ}{\vect{Z}}
\newcommand{\va}{\vect{a}}
\newcommand{\vb}{\vect{b}}
\newcommand{\vc}{\vect{c}}
\newcommand{\vd}{\vect{d}}
\newcommand{\ve}{\vect{e}}
\newcommand{\vf}{\vect{f}}
\newcommand{\vg}{\vect{g}}
\newcommand{\vh}{\vect{h}}
\newcommand{\vi}{\vect{i}}
\newcommand{\vj}{\vect{j}}
\newcommand{\vk}{\vect{k}}
\newcommand{\vl}{\vect{l}}
\newcommand{\vm}{\vect{m}}
\newcommand{\vn}{\vect{n}}
\newcommand{\vo}{\vect{o}}
\newcommand{\vp}{\vect{p}}
\newcommand{\vq}{\vect{q}}
\newcommand{\vr}{\vect{r}}
\newcommand{\vs}{\vect{s}}
\newcommand{\vt}{\vect{t}}
\newcommand{\vu}{\vect{u}}
\newcommand{\vv}{\vect{v}}
\newcommand{\vw}{\vect{w}}
\newcommand{\vx}{\vect{x}}
\newcommand{\vy}{\vect{y}}
\newcommand{\vz}{\vect{z}}
\newcommand{\valpha}{\vectg{\alpha}}
\newcommand{\vbeta}{\vectg{\beta}}
\newcommand{\vgamma}{\vectg{\gamma}}
\newcommand{\vdelta}{\vectg{\delta}}
\newcommand{\vepsilon}{\vectg{\epsilon}}
\newcommand{\vtau}{\vectg{\tau}}
\newcommand{\vmu}{\vectg{\mu}}
\newcommand{\vphi}{\vectg{\phi}}
\newcommand{\vPhi}{\vectg{\Phi}}
\newcommand{\vpi}{\vectg{\pi}}
\newcommand{\vPi}{\vectg{\Pi}}
\newcommand{\vPsi}{\vectg{\Psi}}
\newcommand{\vchi}{\vectg{\chi}}
\newcommand{\vvarphi}{\vectg{\varphi}}
\newcommand{\veta}{\vectg{\eta}}
\newcommand{\viota}{\vectg{\iota}}
\newcommand{\vkappa}{\vectg{\kappa}}
\newcommand{\vlambda}{\vectg{\lambda}}
\newcommand{\vLambda}{\vectg{\Lambda}}
\newcommand{\vnu}{\vectg{\nu}}
\newcommand{\vgo}{\vectg{\o}}
\newcommand{\vvarpi}{\vectg{\varpi}}
\newcommand{\vtheta}{\vectg{\theta}}
\newcommand{\vTheta}{\vectg{\Theta}}
\newcommand{\vvartheta}{\vectg{\vartheta}}
\newcommand{\vrho}{\vectg{\rho}}
\newcommand{\vsigma}{\vectg{\sigma}}
\newcommand{\vSigma}{\vectg{\Sigma}}
\newcommand{\vvarsigma}{\vectg{\varsigma}}
\newcommand{\vupsilon}{\vectg{\upsilon}}
\newcommand{\vomega}{\vectg{\omega}}
\newcommand{\vOmega}{\vectg{\Omega}}
\newcommand{\vxi}{\vectg{\xi}}
\newcommand{\vXi}{\vectg{\Xi}}
\newcommand{\vpsi}{\vectg{\psi}}
\newcommand{\vzeta}{\vectg{\zeta}}
\newcommand{\vzero}{\vect{0}}
\newcommand{\cA}{\mathcal{A}}
\newcommand{\cB}{\mathcal{B}}
\newcommand{\cC}{\mathcal{C}}
\newcommand{\cD}{\mathcal{D}}
\newcommand{\cE}{\mathcal{E}}
\newcommand{\cF}{\mathcal{F}}
\newcommand{\cG}{\mathcal{G}}
\newcommand{\cH}{\mathcal{H}}
\newcommand{\cI}{\mathcal{I}}
\newcommand{\cJ}{\mathcal{J}}
\newcommand{\cK}{\mathcal{K}}
\newcommand{\cL}{\mathcal{L}}
\newcommand{\cM}{\mathcal{M}}
\newcommand{\cN}{\mathcal{N}}
\newcommand{\cO}{\mathcal{O}}
\newcommand{\cP}{\mathcal{P}}
\newcommand{\cQ}{\mathcal{Q}}
\newcommand{\cR}{\mathcal{R}}
\newcommand{\cS}{\mathcal{S}}
\newcommand{\cT}{\mathcal{T}}
\newcommand{\cU}{\mathcal{U}}
\newcommand{\cV}{\mathcal{V}}
\newcommand{\cW}{\mathcal{W}}
\newcommand{\cX}{\mathcal{X}}
\newcommand{\cY}{\mathcal{Y}}
\newcommand{\cZ}{\mathcal{Z}}
\newcommand{\fA}{\mathfrak{A}}
\newcommand{\fB}{\mathfrak{B}}
\newcommand{\fC}{\mathfrak{C}}
\newcommand{\fD}{\mathfrak{D}}
\newcommand{\fE}{\mathfrak{E}}
\newcommand{\fF}{\mathfrak{F}}
\newcommand{\fG}{\mathfrak{G}}
\newcommand{\fH}{\mathfrak{H}}
\newcommand{\fI}{\mathfrak{I}}
\newcommand{\fJ}{\mathfrak{J}}
\newcommand{\fK}{\mathfrak{K}}
\newcommand{\fL}{\mathfrak{L}}
\newcommand{\fM}{\mathfrak{M}}
\newcommand{\fN}{\mathfrak{N}}
\newcommand{\fO}{\mathfrak{O}}
\newcommand{\fP}{\mathfrak{P}}
\newcommand{\fQ}{\mathfrak{Q}}
\newcommand{\fR}{\mathfrak{R}}
\newcommand{\fS}{\mathfrak{S}}
\newcommand{\fT}{\mathfrak{T}}
\newcommand{\fU}{\mathfrak{U}}
\newcommand{\fV}{\mathfrak{V}}
\newcommand{\fW}{\mathfrak{W}}
\newcommand{\fX}{\mathfrak{X}}
\newcommand{\fY}{\mathfrak{Y}}
\newcommand{\fZ}{\mathfrak{Z}}
\newcommand{\fc}{\mathfrak{c}}
\newcommand{\fo}{\mathfrak{o}}

%% file: introduction.tex
\section{Introduction}
Robots operating in domestic environments must solve complex, long-horizon tasks such as preparing a cup of tea, making coffee, or tidying a room, tasks that require coordinating interactions across multiple objects and executing structured action sequences over time. While recent progress in policy learning~\cite{chi2023diffusionpolicy, lee2024behavior} and large-scale demonstration collection~\cite{zhao2023learning, fu2024mobile, chi2024universal, wu2023gello} have improved low-level manipulation, imitation learning still struggles with sample efficiency and compositional generalisation. These challenges are amplified in long-horizon, multi-object settings, where task complexity scales quickly and end-to-end policies trained on long, monolithic trajectories often fail to generalise, requiring an impractical number of demonstrations to capture all task variations.

\begin{figure}[t]
    \centering
    \includegraphics[width=1.0\textwidth]{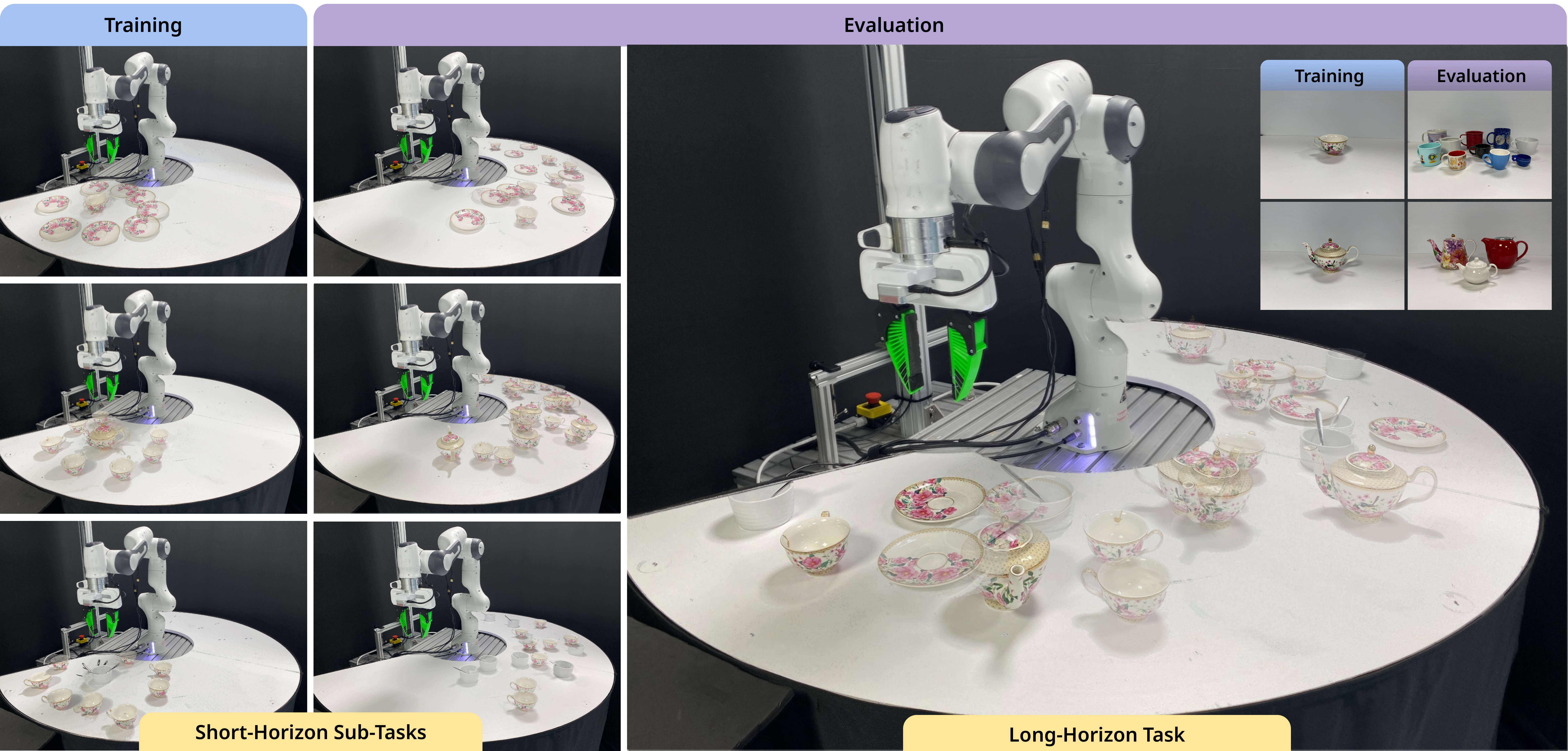}
    \caption{\small{\textbf{Compositional Generalisation.} Our representation allows for independent sub-policy training on task-relevant objects from only 10 demonstrations (left). All sub-policies can then be seamlessly composed at evaluation time to solve long-horizon tasks across a vast range of unseen intra-object spatial configurations (right) where all objects are present in the scene as well as intra-category variations from the objects used during training (top-right).}}
\vspace{-0.5cm}
    \label{fig:start-config}
\end{figure}

A promising approach to addressing such tasks is to simplify the learning problem towards learning sub-policies that can be composed to solve the longer-horizon task \cite{uvd, gupta2019relay, heo2023furniturebench, lee2021ikea, mandlekar2020learning}. This however presents several challenges: 1) identifying how to partition the task into sub-tasks that can be independently learned; 2) the distribution shift encountered by sub-policies when presented with the full task configuration and 3) the need to learn an additional arbitration policy that knows when to switch between each sub-policy. 

In this work, we take an affordance-centric perspective to address each of these limitations. First, we partition long-horizon, multi-object tasks into affordance-aligned sub-tasks, each defined around a localised object interaction, such as pouring from a teapot or grasping a cup. This provides a natural and task-relevant decomposition that enables independent training of sub-policies focused on functionally distinct interactions.

Second, we address the distribution shift that arises when composing sub-policies in long-horizon tasks, a challenge exacerbated by two dominant policy representations: image-based inputs~\cite{chi2023diffusionpolicy, padalkar2023open, lee2024behavior, bharadhwaj2024roboagent} and global coordinate frames~\cite{bharadhwaj2024roboagent, padalkar2023open, ceola2024robotic, brohan2022rt, fang2023rh20t, mandlekar2018roboturk}. Image-based policies trained in isolation fail to capture the full range of visual variations encountered when all task-relevant objects appear at test time, while global frames require demonstrations to cover all possible spatial configurations. To address this, we introduce oriented affordance frames: object-centric coordinate systems anchored at task-relevant affordances and oriented toward the robot’s tool frame. These frames retain only the functionally relevant structure of the task, abstracting away clutter and irrelevant details. By rotating the frame with respect to the tool, each sub-policy is trained in a consistent local reference frame, ensuring it remains in-distribution even under novel robot start configurations encountered during policy composition. Grounding policies in these relative frames supports generalisation to spatial variations and novel arrangements without requiring exhaustive demonstration coverage, and naturally enables intra-category generalisation to objects with different appearances or geometries ~\cite{ke2021grasping, di2024dinobot, huang2024rekep, pan2025omnimanip}. While affordance representations have been explored in prior work~\cite{manuelli2019kpam, di2024dinobot, simeonovdu2021ndf, wang2024d3fields}, their use in closed-loop behaviour cloning for robust, composable policy learning remains under-explored. 

Third, we augment each sub-policy with a continuous self-progress prediction signal, learned directly from the length of demonstration trajectories. This scalar output allows the system to autonomously transition between sub-policies without requiring a separate arbitration policy or external supervision, enabling smooth and robust policy composition across extended task horizons.

Our work makes three key contributions toward scalable and generalisable imitation learning for long-horizon, multi-object manipulation tasks. (1) We introduce the concept of the oriented affordance frame, a local, task-aligned reference frame that enables sub-policy learning to be both spatially invariant and compositionally robust. (2) We develop a perception pipeline that leverages pre-trained vision foundation models to detect and track these affordance frames without reliance on fiducial markers, supporting real-world deployment. (3) We augment each sub-policy with a continuous self-progress prediction signal, enabling automatic and reliable arbitration between sub-tasks without requiring a high-level controller. 

Through real-world experiments, we show that our affordance-centric approach enables sample-efficient policy learning from just 10 demonstrations per sub-task, while significantly outperforming image-based and global-frame baselines. It generalises robustly to unseen spatial configurations and novel object instances, and supports seamless composition of independently trained sub-policies. Additionally, our marker-free perception pipeline maintains high performance, demonstrating the practicality of our approach in realistic settings.

%% file: related_work.tex
\section{Related Work}

\textbf{Generalisation in Behaviour Cloning:} Recent advances in generative modelling have revived interest in behaviour cloning for learning complex, multi-modal behaviours from demonstrations~\cite{chi2023diffusionpolicy, fu2024mobile, zhao2023learning, lee2024behavior}. Behaviour cloning typically maps input states often images or point clouds due to their generality and ease of collection~\cite{chi2023diffusionpolicy, padalkar2023open, lee2024behavior, bharadhwaj2024roboagent} to actions. A major challenge is covariate shift, where small differences between training and test inputs, especially in high-dimensional image spaces, can degrade performance~\cite{ross2011reduction, torne2024reconciling}. Current efforts in generalisation focus on large-scale data collection~\cite{bharadhwaj2024roboagent, brohan2022rt, fang2023rh20t, padalkar2023open} or architectural invariances~\cite{weiler2019general, wang2022equivariant, zhu2022sample}. However, these approaches mainly tackle spatial generalisation and often retain task-irrelevant details, limiting policy compositionality. We instead propose learning affordance-centric 3D task frames that discard irrelevant information and enable robust intra-category, spatial, and compositional generalisation.

\textbf{Keypoint-based Representations for Manipulation:} Keypoints have been widely used in robotic manipulation to enable intra-category generalisation by focusing on task-relevant object regions~\cite{manuelli2019kpam, breyer2021volumetric, jiang2021synergies, kulkarni2019unsupervised, florence2018dense, simeonovdu2021ndf, wang2024d3fields, task-axes}. Early approaches trained custom vision models to detect keypoints and solved task-specific SE(3) optimisations for single-step, open-loop tasks~\cite{manuelli2019kpam, florence2018dense}. More recent methods~\cite{dipalo2024kat, wang2024d3fields} leverage pre-trained vision models to extract keypoints or segmentations~\cite{zhu2022viola}, reducing the need for task-specific training but still operating in open-loop settings with limited spatial invariance. In contrast, we focus on closed-loop behaviour cloning, using keypoint regions as 3D task frames to achieve both spatial and intra-category invariance. We additionally propose a general, task-agnostic pipeline that leverages foundation models to extract keypoint regions without training custom models.

\begin{figure}[t]
    \centering
    \includegraphics[width=1.0\textwidth]{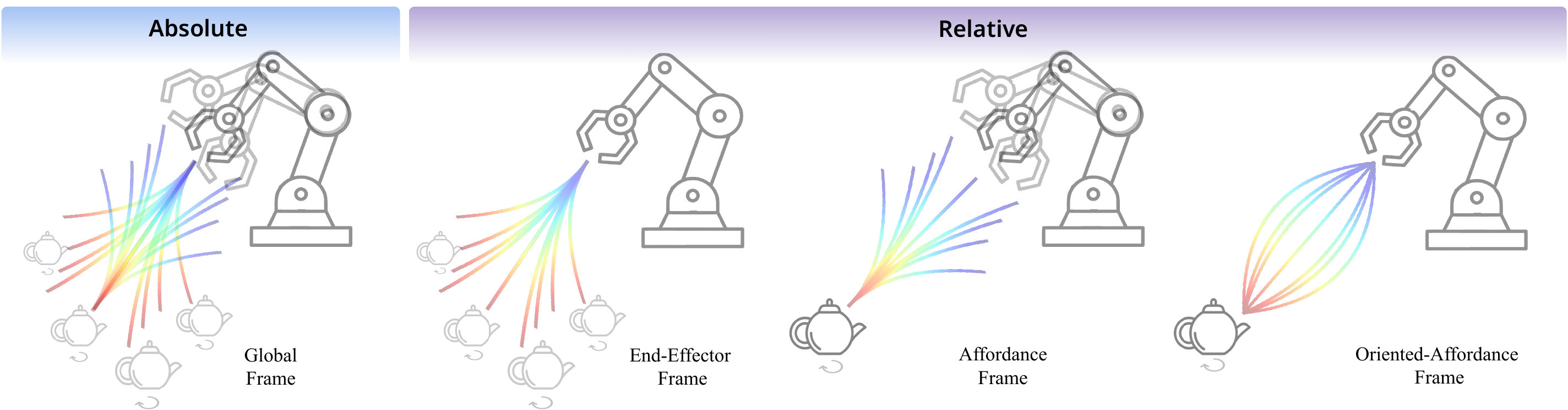}
    \caption{\small{\textbf{Comparison of different reference frames for policy learning.} A global reference frame (left) requires demonstrations covering all spatial variations of both the object and the end-effector, leading to poor spatial generalisation. End-effector and affordance-centric frames (middle) reduce this requirement but still require extensive data to capture relative transformations. The proposed oriented affordance frame (right) aligns the state and action representation with the task-relevant affordance and tool frames, thus ensuring spatial invariance while minimising data requirements for policy learning.}
    \vspace{-0.45cm}
    }
    \label{fig:reference_frames}
\end{figure}

\textbf{Task Frames:} Task frames have long been used in classical robotics to define motions relative to objects or tools to simplify motion generation~\cite{ballard1984task, raibert1981hybrid, mason1981compliance, berenson2011task, king2016rearrangement, migimatsu2020object}. Recent works have adapted this idea to reinforcement learning and behaviour cloning. \citet{chi2024universal} introduced an end-effector–based task frame to simplify in-the-wild data collection, but it still relied on image state representations, which lack task-centric invariances, requiring large-scale demonstrations to generalise. \citet{ke2021grasping} improved spatial invariance by attaching frames to object centres, preserving relative transformations and improving data efficiency for simple tasks. However, their method did not account for object rotations or intra-category variations, limiting generalisation with a tendency to violate robot kinematics when the object rotated beyond certain limits. We address these limitations with a simple approach that supports arbitrary object orientations and generalisation across object instances by introducing an oriented affordance task frame for behaviour cloning.

%% file: method.tex
\section{Affordance-Centric Policy Learning with Oriented Affordance Frames}

The goal of our work is to train state-conditioned robotic policies $\pi(\va_t|\vs_t)$ that are 1) sample-efficient, i.e. they can be learned from as little as 10 human demonstrations; 2) invariant to the spatial configuration of the task-relevant objects; 3) invariant to variations in the object geometry and appearance; and 4) composable to solve multi-step and long-horizon tasks involving multiple interacting objects.    

To achieve this goal, we replace the image- or point cloud- based state representation that is currently prevalent in many imitation learning approaches ~\cite{chi2023diffusionpolicy, padalkar2023open, lee2024behavior, bharadhwaj2024roboagent, 3d_diffuser_actor, Ze2024DP3}. Instead, we represent the state $\vs_t$ as the pose of the currently task-relevant \emph{tool frame} relative to an \emph{oriented affordance frame}. The latter is a reference frame that is centred on the currently relevant affordance \emph{and} oriented towards the origin of the tool frame at the start of the task. 
We will describe these coordinate frames in detail in the following (Sec.~\ref{sec:frames}), before introducing a perception pipeline that can automatically detect and track these frames on objects unseen during training (Sec.~\ref{sec:perception}). In Section~\ref{sec:progress}, we will describe our proposed method of policy arbitration that enables the autonomous composition of multiple policies to solve long-horizon and multi-step tasks.

\subsection{Oriented Affordance Frames for States and Actions}
\label{sec:frames}
The choice of reference frame for representing state and actions significantly affects a policy's ability to generalise to spatial variations in multi-object tasks, as illustrated in Fig.~\ref{fig:reference_frames}. When states and actions are represented in a fixed global reference frame (first panel), demonstration trajectories must densely cover variations in object and robot poses to attain spatial generalisation. If an object appears in a previously unseen global position, the policy will be out of distribution and likely fail. 

Using relative coordinate frames, e.g. expressing the robot's actions relative to its current end-effector or using the pose of task-relevant objects relative to the end-effector, are simple examples of using \emph{relative} reference frames. As illustrated in the middle panels of Fig.~\ref{fig:reference_frames}, these partly alleviate generalisation problems but still require demonstrations to cover all possible poses of task-relevant objects relative to the robot (2\textsuperscript{nd} panel in Fig.~\ref{fig:reference_frames}), or vice-versa (3\textsuperscript{rd} panel), to avoid the policy being out-of-distribution when encountering a previously unseen pose of the object relative to the robot.

\begin{wrapfigure}{l}{0.6\columnwidth}
  \vspace{-7pt} 
  \includegraphics[width=0.6\columnwidth]{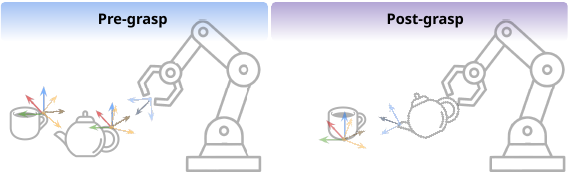}
  \caption{\small{\textbf{Affordance Frames, Oriented-Affordance Frames and Tool Frames.} Left: Affordance frames (\raisebox{-0.2em}{\includegraphics[height=1.2em]{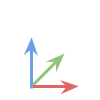}}), oriented affordance frames (\raisebox{-0.2em}{\includegraphics[height=1.2em]{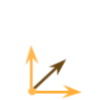}}), and tool frame (\raisebox{-0.2em}{\includegraphics[height=1.2em]{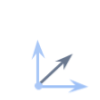}}) for a typical \emph{pick} task. Right: Frames for the \emph{pour} task. The oriented affordance frames have the same origin as the affordance frames, but are oriented such that one of the axes (brown) points towards the origin of the tool frame at the beginning of the task.}}
  \label{fig:frames_explained}
\end{wrapfigure}


\textbf{Affordance Frames:} 
Relative reference frames can not only be centred on the robot's end-effector but also on the affordances of task-relevant objects. Objects can have multiple task-dependent localised \emph{affordances}: e.g. a cup has an affordance on the handle for the task of \emph{picking up} and an affordance in the centre of the cup for the task of \emph{pouring}. Our work makes extensive use of affordance frames, but -- importantly and in contrast to previous work~\cite{gao2023k, gao2021kpam} -- orients them towards the current tool frame and leverage these representations in the behaviour cloning setting.

\textbf{Tool Frames:} In multi-object tasks, a robot either directly interacts with an object (e.g. when picking it up, pushing or opening it), or acts on a target object while holding a \emph{tool} object, e.g. a spoon. In addition to the affordance frame defined on to the \emph{target} object, we define a \textit{tool frame} on the \emph{tool object}. For simple pickup tasks, the tool frame is identical to the robot's end-effector frame, however for actions such as stirring tea with a spoon or pouring from a teapot, the tool frame is placed on the scoop of the spoon or the spout of the teapot respectively as illustrated in Fig.~\ref{fig:frames_explained}.

\textbf{Oriented-Affordance Frames: } Given affordance frames and tool frames, we can now introduce the \textit{oriented-affordance frame}, a core concept of our paper. The oriented affordance frame is obtained by rotating the affordance frame on the target object such that one of its axes (we consistently choose the x-axis as this ``funnel'' axis) is directed towards the origin of the tool frame. 

The oriented affordance frame is the central concept for our generalisable and sample-efficient policy learning: we represent both the state $\vs_t$ and the action sequences $\va_t$ of our trained policies $\pi(\va_t|\vs_t)$ in the oriented affordance frame.

\textbf{Frame Initialisation and Update: }
We initialise the oriented affordance frame at the start of each task. Our perception pipeline (detailed in Appendix ~\ref{sec:perception}) extracts the pose of the currently relevant affordance frame ${}^\text{W}\vT_\text{afford}\in \vS\vE(3)$ in a global world reference frame. With knowledge of the forward kinematics of the robot and the currently held tool object (if any), we also know the pose of the tool frame ${}^\text{W}\vT_\text{tool}$. Using Algorithm~\ref{algorithm1} we calculate the rotation matrix $\vR_\text{align}$ that transforms the affordance frame such that its x-axis points towards the origin of the tool frame, thus yielding the oriented affordance frame ${}^\text{W}\vT_\text{o-aff} = \vR_\text{align}\cdot {}^\text{W}\vT_\text{afford}$.

While the origin of the oriented affordance frame can move during task execution if the robot moves the target object, its \emph{orientation} is kept anchored so that the x-axis keeps pointing to where the origin of the tool frame was \emph{at the beginning of the task}. Our experimental ablation in Sec.~\ref{sec:results} will demonstrate the benefit of this small but important detail.

\textbf{State Representation: }
The state $\vs_t$ for our policy comprises the current pose of the tool-frame in the oriented affordance frame ${}^{\text{o-aff}}\vT_{\text{tool}} \in \vS\vE(3)$, the binary gripper state $g_{\text{s}} \in \{0, 1\}$, and the rotation ${}^\text{o-aff}\vR_\text{aff}$ of the target object relative to its anchored oriented-affordance frame.

\textbf{Action Representation: }
The actions generated by the policy consist of a sequence of $N=16$ desired next poses of the robot's end effector in the oriented affordance frame $\{{}^{\text{o-aff}}\vT_{\text{ee}}\}_{\tau=t\dots t+N}$ and a sequence of gripper actions $g_{\text{a}} \in \{0, 1\}$ of equal length $N$. 

\begin{wrapfigure}{l}{0.4\columnwidth}
  \vspace{-7pt} 
  \includegraphics[width=0.9\linewidth]{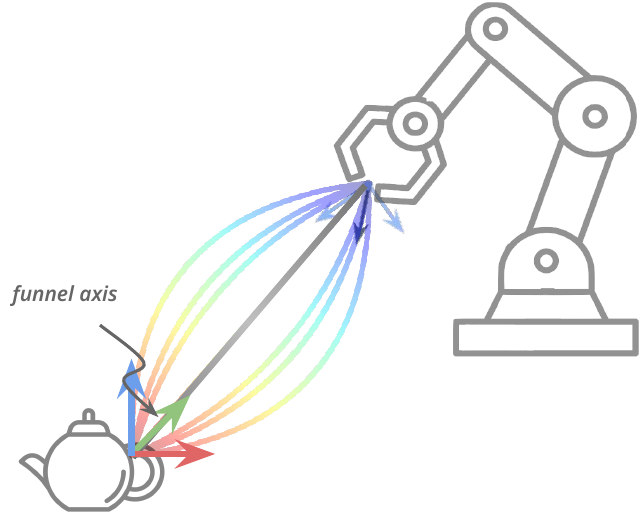}
  \caption{\small{\textbf{Adaptive Data Support of the Oriented Affordance Frame.} By aligning the affordance frame with the tool frame at task initiation, demonstrated trajectories become aligned along a consistent `funnel' axis, reducing variability and improving generalisation. This ensures that policies trained in the oriented affordance frame remain robust to changes in object rotation and robot poses, facilitating seamless policy composition.}}
  \label{fig:funnel}
  \vspace{-30pt} 
\end{wrapfigure}

\textbf{Intuitive Benefits of the Oriented Affordance Frame: }
Expressing robot state and actions relative to the oriented affordance frame maximises the utility of a small number of demonstrations. Intuitively, all demonstrated trajectories tend to be aligned or, to some degree, in the vicinity of the oriented (``funnel'') axis, independent of the relative poses of the robot and target object, as illustrated in Figure~\ref{fig:funnel}. 
When composing multiple policies, the oriented affordance frame representation ensures that the robot's tool frame at the end of a task is always \emph{in distribution}, i.e.\ within the data support of the following policy regardless of the end-effector's global location or the absolute pose of the target object. 

\subsection{Policy Arbitration by Self-Progress Prediction}
\label{sec:progress}
With the appropriate abstractions in place, we can now train independent, affordance-centric sub-policies that can be composed to solve longer-horizon tasks. To support automatic policy composition and arbitration, we augment the action space and add a scalar policy \emph{self-progress indicator} $a_\text{progress} \in [0,1]$. During policy training, we compute a task progress measure for a demonstration trajectory by linearly interpolating from 0 to 1 based on the duration of the trajectory. The policy is then trained to output actions and the corresponding progress value in the added progress indicator $a_\text{progress}$. During policy execution, this self-progress estimate determines when to transition from one sub-policy to the next, based on a simple threshold. This lets us compose sub-policies to solve complex long-horizon tasks without training an additional arbitration policy. 

%% file: evaluation.tex
\section{Evaluation}
We describe the extensive experiments conducted to support the key claims our paper makes regarding (i) sample-efficient policy training from as little as 10 demonstrations, achieving substantially better performance than other representations; (ii) spatial generalisation; (iii) generalisation to new objects unseen during training; and (iv) the automatic arbitration between sub-policies in long-horizon, multi-step tasks.

We focused our experiments on three multi-step, multi-object tasks representative of scenarios that future domestic service robots are likely to encounter. The first task, preparing a cup of tea, is the primary focus for our quantitative analysis. Additionally, two supplementary tasks -- making coffee and putting a pair of shoes on a rack -- are included for qualitative evaluation. Videos demonstrating the trained policies autonomously executing all tasks are available on the anonymous project page: \href{https://affordance-policy.github.io/}{https://affordance-policy.github.io/}.

\subsection{Experimental Setup -- Tea Making}

\label{sec:experimental_setup}

\begin{figure*}[t]
    \centering
    \includegraphics[width=1.0\textwidth]{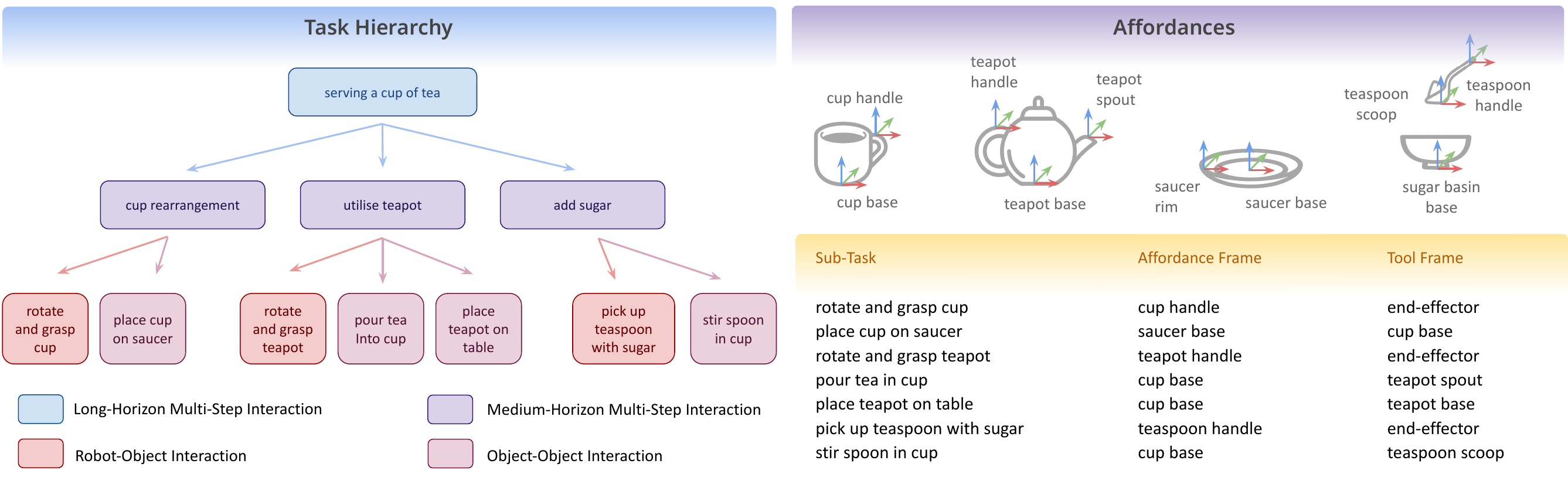}
    \caption{\small{\textbf{Affordance-centric task decomposition for the tea serving task.} \textit{Left:} Task decomposition hierarchy; \textit{Top Right:} Affordance-centric frames for each object; \textit{Bottom Right:} Sub-task frame definitions.}}
    \vspace{-0.4cm}
    \label{fig:task-hierarchy}
\end{figure*}

\textbf{Task Description: } 
We focus our quantitative analysis on a tea-serving task involving five objects: teacup, saucer, teapot, sugar bowl, and teaspoon. The task is decomposed into seven sequential sub-tasks, each defined by the target object and relevant affordance: (1) rotating and grasping the teacup; (2) placing the cup on the saucer; (3) rotating and grasping the teapot; (4) pouring tea into the cup; (5) placing the teapot on the table; (6) scooping sugar with the spoon; and (7) transferring sugar into the cup, stirring, and placing the spoon on the saucer.

We evaluate our method and baselines across these sub-tasks, their compositions (e.g., combining tasks 3–5 into \emph{utilise teapot}), and the full tea-serving sequence, as shown in Fig.\ref{fig:task-hierarchy}. For each object, we define the corresponding affordance and tool frames (Fig.\ref{fig:task-hierarchy}). As noted in Appendix~\ref{sec:limitations}, we assume the human demonstrator can identify affordances and meaningful task partitions.

\textbf{Perception System:}
We evaluate our method using two perception setups, explicitly indicated for each experiment (Sec.\ref{sec:results}):\\
\textit{1) Marker-based:} To isolate the impact of the oriented affordance frames from perception performance (Appendix\ref{sec:perception}), we use ArUco markers to obtain ground-truth affordance poses.\\
\textit{2) Large Vision Models:} A subset of experiments—including those shown in the supplementary videos—employ our proposed perception pipeline based on pre-trained vision models (Appendix~\ref{sec:perception}, Fig.~\ref{fig:main}).


\textbf{Policy Training:} We use Diffusion Policy~\cite{chi2023diffusionpolicy} for imitation learning, training each policy for 4500 epochs with the default parameters from the original implementation. The 16-dimensional state space includes the robot's tool frame pose relative to the oriented affordance frame ${}^{\text{o-aff}}\vT_{\text{tool}}$, the binary gripper state $g_{\text{s}} \in {0, 1}$, and the object's orientation relative to the oriented affordance frame ${}^\text{o-aff}\vR_\text{aff}$. Both ${}^\text{o-aff}\vR_\text{aff}$ and the rotation component of ${}^{\text{o-aff}}\vT_{\text{tool}}$ are expressed as 6D vectors, following~\cite{zhou2019continuity}. For baselines, we additionally provide the 3D position of the current target object.

The action space is 11-dimensional, comprising the 3D robot position, 6D robot orientation \cite{zhou2019continuity}, the 1D gripper action, and the 1D self-progress prediction. Our method represents the end-effector pose in the oriented affordance frame, while baselines use either the end-effector frame, the affordance frame or the global frame. Following Diffusion Policy's temporal action generation, the policy outputs a sequence of 16 actions, resulting in a 176-dimensional output vector.

\subsection{Results}
We report the results of our key experiments and ablation studies below. Each set of experiments supports one of the core claims of our paper.

\label{sec:results}
\begin{table}[t]
\centering
\caption{\small{\textbf{Summary of Results.} Success rates for in-distribution (IND) and out-of-distribution (OOD) scenarios for various tasks and composite tasks.}}
\resizebox{\textwidth}{!}{%
\begin{tabular}{@{}rcc>{\columncolor[gray]{.9}[0.2pt][0.2pt]}cc>{\columncolor[gray]{.9}[0.2pt][0.2pt]}cc>{\columncolor[gray]{.9}[0.2pt][0.2pt]}c@{}}
\toprule
& & \multicolumn{2}{c}{\textbf{\begin{tabular}[c]{@{}c@{}}Oriented Affordance \\ Frame (Ours)\end{tabular}}} & \multicolumn{2}{c}{\textbf{End Effector Frame}} & \multicolumn{2}{c}{\textbf{Global Frame}} \\
\cmidrule(lr){3-4} \cmidrule(lr){5-6} \cmidrule(lr){7-8}
\textbf{Task} & \textbf{Demos} & \textbf{IND Success} & \textbf{OOD Success} & \textbf{IND Success} & \textbf{OOD Success} & \textbf{IND Success} & \textbf{OOD Success}\\
\midrule
\multicolumn{8}{>{\columncolor[gray]{0.9}}l}{\textbf{Base Tasks}} \\
(T1) rotate and grasp cup & 10 & 81.8\% & 81.8\% & 45.5\% & 45.5\% & 45.5\% & 0.0\%\\
(T2) place cup on saucer & 10 & 100\% & 100\% & 100\% & 100\% & 9.1\% & 0.0\% \\
(T3) rotate and grasp teapot & 10 & 90.9\% & 81.8\% & 27.3\% & 27.3\% & 81.8\% & 0.0\% \\
(T4) pour tea into cup & 10 & 100\% & 81.8\% & 45.5\% & 27.3\% & 54.5\% & 0.0\%\\
(T5) place teapot on table & 10 & 90.9\% & 72.7\% & 54.5\% & 54.5\% & 90.9\% & 0.0\%\\
(T6) pick up teaspoon with sugar & 10 & 81.8\% & 81.8\% & 45.5\% & 27.3\% & 72.7\% & 0.0\%\\
(T7) stir spoon in cup & 10 & 90.9\% & 81.8\%& 18.2\% & 9.1\% & 72.7\% & 0.0\% \\
\textbf{Average} & 10 & \textbf{90.9\%} & \textbf{83.1\% }& 48.1\% & 41.6\% & 59.7\% & 0.0\%\\
\midrule
\multicolumn{8}{>{\columncolor[gray]{0.9}}l}{\textbf{Composite Tasks}} \\
(T1+T2) cup rearrangement & 10 & 81.8\% & 81.8\% & 45.5\% & 0.0\% & 36.4\% & 0.0\% \\
(T3+T4+T5) utilise teapot & 10 & 81.8\% & 63.6\% & 18.2\% & 9.1\% & 45.5\% & 0.0\% \\
(T6+T7) add sugar & 10 & 72.2\% & 72.2\% & 9.1\% & 9.1\% & 63.6\% & 0.0\% \\
\textbf{Average} & 10 & \textbf{78.8\%} &\textbf{ 72.7\%} & 24.2\% & 6.1\% & 48.5\% & 0.0\% \\
\midrule
\multicolumn{8}{>{\columncolor[gray]{0.9}}l}{\textbf{Complete Task}} \\
(T1+T2+T3+T4+T5+T6+T7) serve tea & 10 & \textbf{81.8\%} & \textbf{63.6\%} & 9.1\% & 9.1\% & 0.0\% & 0.0\% \\
\bottomrule
\end{tabular}
}
\label{tab:sub-policy-eval}
\end{table}

\begin{figure*}[t]
    \centering
    \includegraphics[width=1.0\textwidth]{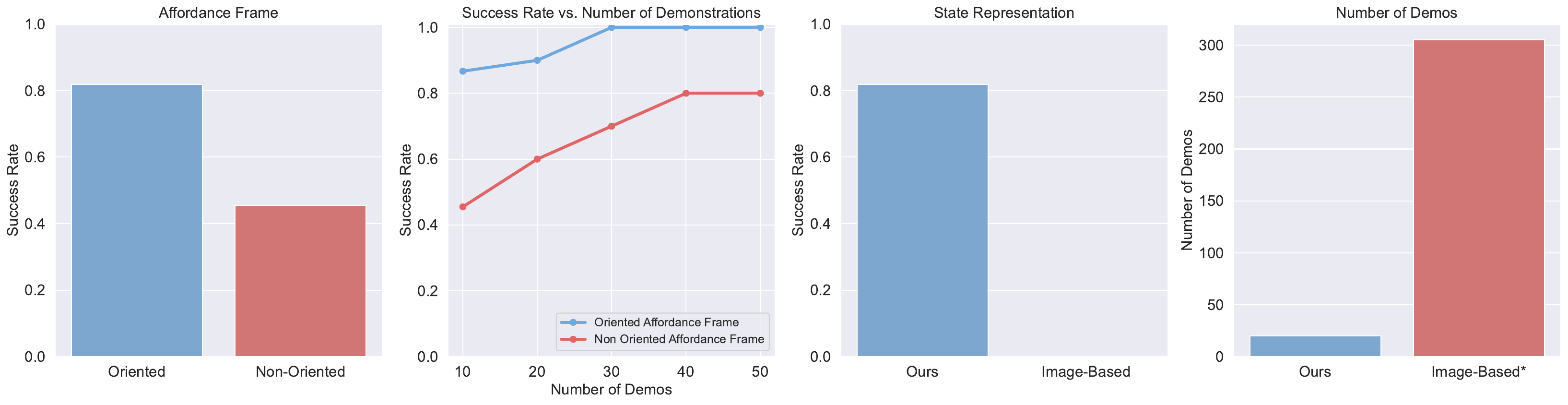}
    \caption{\small{\textbf{Additional Comparisons.} \textit{a)} Comparison on different affordance frames; \textit{b)} Success rate vs. number of demonstrations for the different affordance frames; \textit{c)} Performance of an image-based RGB policy when trained with only 10 demonstrations for the cup rearrangement task; \textit{d)} Relative number of demonstrations required for standard image-based diffusion policy \citep{chi2024universal} to achieve the same generalisation and performance as our system. }}
    \label{fig:ablation}
    \vspace{-0.4cm}
\end{figure*}

\textbf{Sample-efficient Policy Learning from Only 10 Demonstrations: } Our first experiment demonstrates that the proposed oriented affordance frame enables highly sample-efficient policy learning from just 10 demonstrations. We evaluated this on seven tasks from the tea-serving scenario, along with their composite sequences (Fig.~\ref{fig:task-hierarchy}), using ArUco markers to obtain ground-truth poses for the affordance frames. As shown in Table~\ref{tab:sub-policy-eval}, our method achieved a 90.9\% average success rate across all seven tasks, outperforming the end-effector (48.1\%) and global (59.7\%) frame baselines. This performance advantage persisted across composite tasks, with our method also exceeding 80\% success on the full tea-serving sequence, comprising all seven subtasks in order.

We further examined the \emph{cup rearrangement} task, which combines rotating and grasping the cup followed by placing it on the saucer. Fig.~\ref{fig:ablation} shows that an image-based policy similar to~\cite{chi2024universal} failed completely with 10 demonstrations, requiring 305 demonstrations to match our 81.8\% success—representing a $30\times$ increase in data requirements. Finally, we ablated the orientation component of our frame representation. Replacing ${}^\text{o-aff}\vT_\text{tool}$ with a non-oriented version (${^\text{afford}\vT_\text{tool}}$) nearly halved success rates at 10 demos and failed to surpass 80\% even with 50. In contrast, our method reached 100\% success with just 30 demonstrations, highlighting the critical role of the oriented frame in maximising the utility of a small number of demonstrations.


\textbf{Spatial Generalisation: }
Our second experiment shows that training with the oriented affordance frame representation leads to better spatial generalisation than other representations, especially when training from a few demonstrations. While the experiment described above evaluated the learned policies under in-distribution conditions (denoted IND in Table~\ref{tab:sub-policy-eval}) where the objects were placed in the same sector of the robot's workspace during training and evaluation, we now vary the spatial configurations and place the objects in different parts of the robot workspace during evaluation. See Fig.~\ref{fig:start-config} for a visualisation. We again use the fiducial markers to provide ground-truth poses of the affordance frames. Under these Out-of-Distribution (OOD) conditions, our proposed representation again performs the best, achieving 83.1\% success on average across all base seven tasks, 72.7\% on the composite tasks, and still 63.6\% on the overall tea-serving task that composes all seven task. The end-effector-centric representation performs much worse (41.6\% on average for the base tasks, 6.1\% for the compositions and 9.1\% for the complete tea-serving scenario) and representing state and action in a global frame fails completely throughout all experiments. A detailed breakdown is provided in Table~\ref{tab:sub-policy-eval}.




\begin{wrapfigure}{r} {0.4\textwidth}
\vspace{-0.5cm}
    \centering
    \includegraphics[width=0.35\columnwidth]{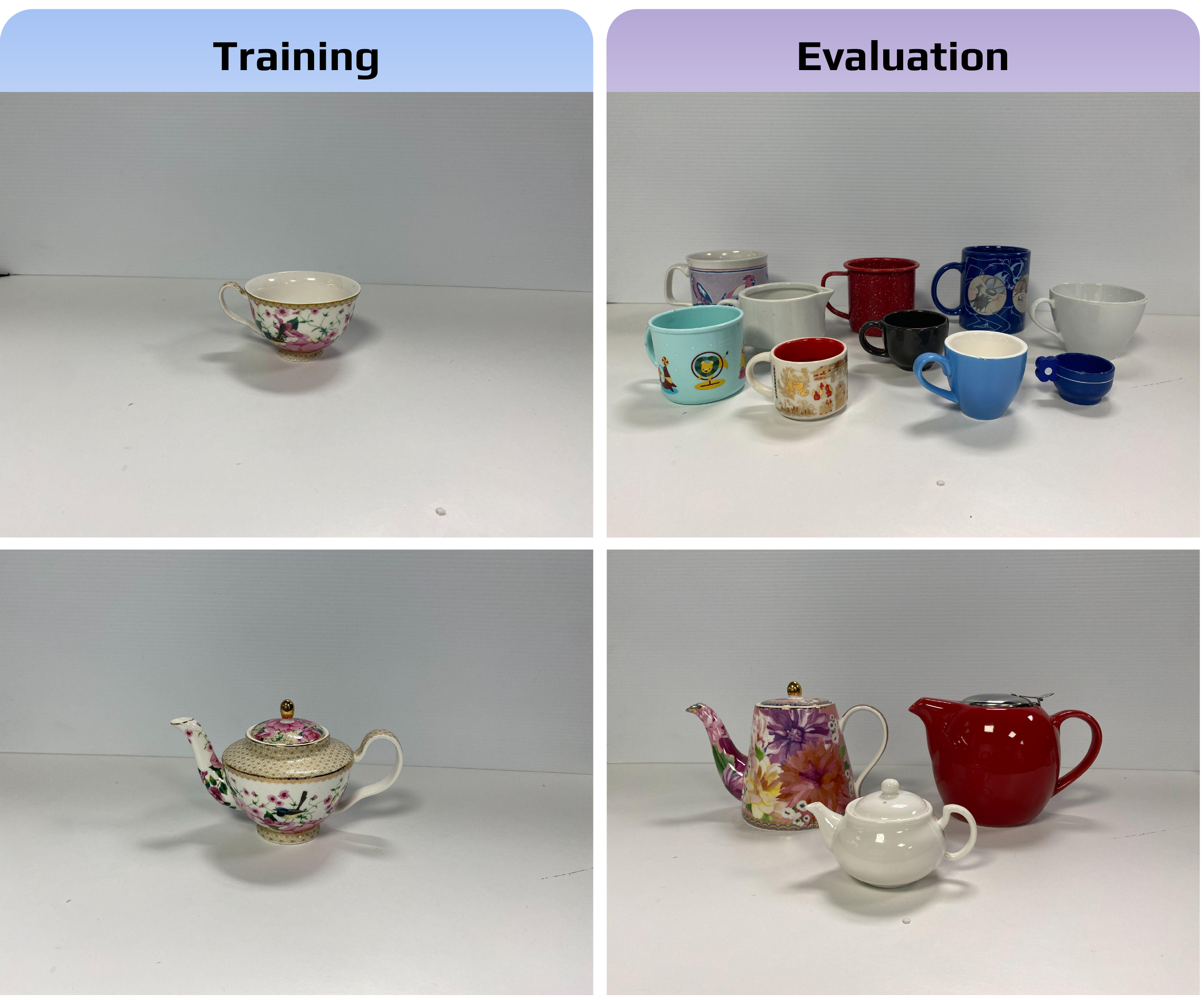}
    \resizebox{0.35\textwidth}{!}{%
\setlength{\tabcolsep}{0.75\tabcolsep}
\begin{tabular}{@{}lcc@{}}
        \toprule
        \multicolumn{1}{c}{\textbf{Task}} & \textbf{\begin{tabular}[c]{@{}c@{}}\# Instances\end{tabular}} & \textbf{Success} \\ \midrule
        cup rearrange                 & 10                                                                                 & 8/10            \\
        utilise teapot                & 3                                                                                  & 3/3              \\ \bottomrule\\
        \end{tabular}%
        }
\caption{\small{\textbf{Generalisation to intra-category variations} The set of objects used for training and evaluating the intra-category generalisation capabilities of the trained sub-policies.}}
\vspace{-0.6cm}
    \label{fig:intra-cat}
\end{wrapfigure}

\textbf{Intra-Category Generalisation: }
Our fourth set of experiments supports our claim that oriented affordance frames enable the transfer of trained policies to new objects unseen during training. These experiments use the perception system from~\ref{sec:perception}.

\begin{figure*}[t]
    \centering
    \includegraphics[width=1.0\textwidth]{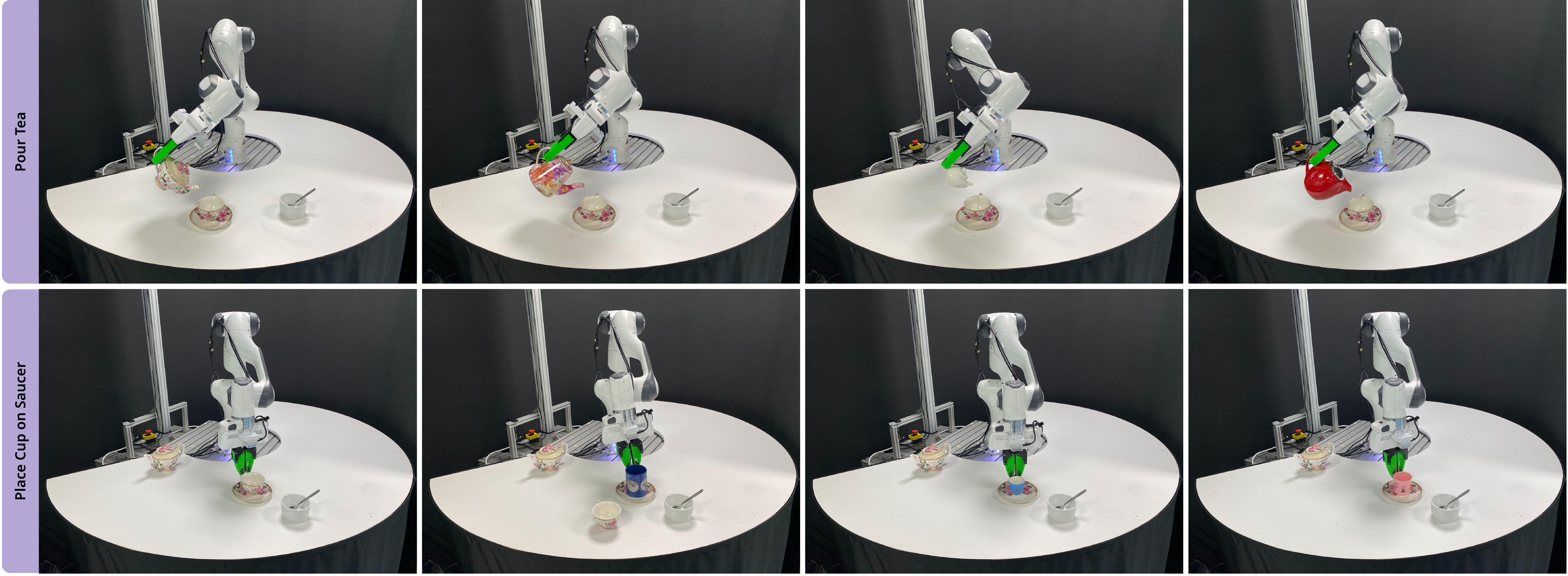}
    \caption{\small{\textbf{Intra-category generalisation.} We demonstrate the ability of our approach to enable generalisation across large shape and size intra-category variations.
    }}
    \vspace{-0.5cm}
    \label{fig:appendix_intra_cat}
\end{figure*}

As illustrated in Fig.~\ref{fig:intra-cat}, we collected all demonstrations for the \emph{cup rearrange} and \emph{utilise teapot} composite tasks with a single teapot and teacup. The learned policies successfully executed on 8 of the 10 unseen teacups and all 3 unseen teapots. These new objects vary significantly in appearance and geometry compared to the demonstration objects. This successful intra-category generalisation is possible due to the proposed perception pipeline's ability to identify and transfer affordances from one object to another despite considerable intra-category variations in geometry and appearance. See Fig.~\ref{fig:appendix_intra_cat} for further illustration of the generalisation in action.




\textbf{Task Composition and Self-Progress Prediction: }
Throughout our experiments involving composite tasks (e.g. \emph{cup rearrangement} or the full tea-making task), we utilise the proposed self-progress prediction mechanism to fully autonomously control the transition between base tasks. We found the self-progress prediction to be remarkably robust in all experiments and did not observe it to cause any failures. More detailed results are provided in Appendix \ref{app:policy_comp}.

%% file: discussion.tex
\section{Conclusion}
\label{sec:discussion}

Our experiments demonstrate that oriented affordance frames substantially enhance both sample efficiency and generalisation in imitation learning for long-horizon, multi-object tasks. By replacing dense image- or point-cloud-based representations with an abstracted, affordance-centric formulation, our approach enables robust policy learning from as few as 10 demonstrations. It generalises effectively across spatial, intra-category, and combinatorial variations - crucial for solving complex, long-horizon tasks. We also introduce a perception pipeline to detect and track these frames using vision foundation models, along with a simple yet effective progress estimation metric derived from demonstration duration to enable seamless sub-policy transitions. We hope this work encourages further exploration of structured representations, priors, and compositionality in behaviour cloning, paving the way toward more generalisable and practical robotic systems for real-world applications.

\section*{Acknowledgments}
The authors also acknowledge the ongoing support from the QUT Centre for Robotics. This work was partially supported by the Australian Government through the Australian Research Council's Discovery Projects funding scheme (Project DP220102398) and by an Amazon Research Award to Niko S\"underhauf. This work was also supported by the QUT Research Engineering Facility. 

%% file: appendix.tex
\clearpage
\appendix

\section{Appendix}

\subsection{A Perception Pipeline to Detect and Track Affordance Frames}
\label{sec:perception}

\begin{figure*}[h]
  \centering
  \includegraphics[width=\textwidth]{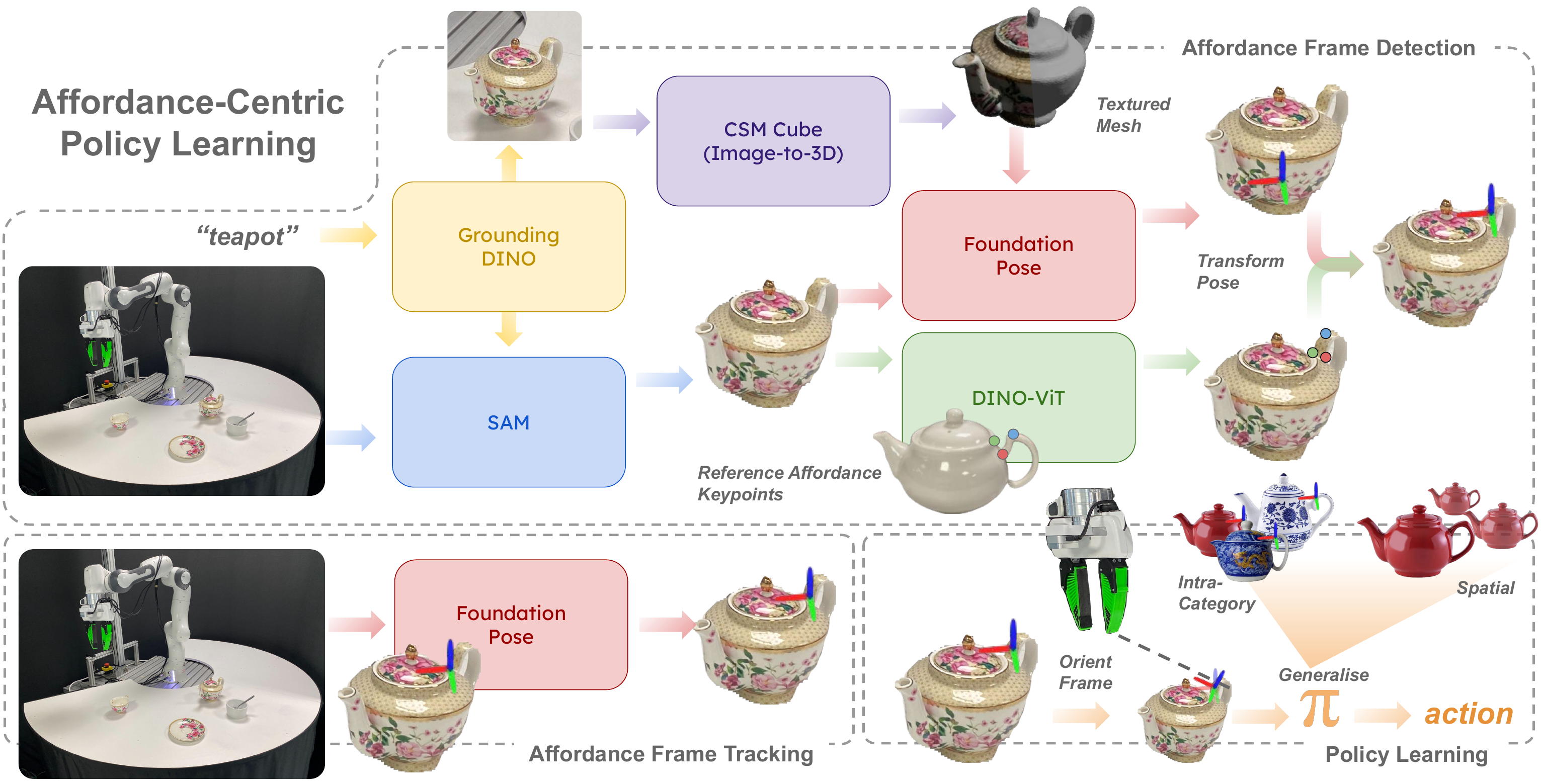}
  \caption{\small{\textbf{Affordance-Centric Policy Learning.} \textit{Affordance Frame Detection:} We propose a framework to detect affordance frames using pre-trained large vision models. \textit{Affordance Tracking:} Once the frame is detected we utilise Foundation Pose~\cite{wen2023foundationpose} to continuously track the frame in real-time as the robot interacts with it. \textit{Policy Learning:} At the start of each episode we appropriately orient the frame towards the tool frame of the robot and train a state-based diffusion policy that operates with this frame as its task frame.}}
  \label{fig:main}
\end{figure*}

Automatically detecting and tracking the affordance frames necessary for the affordance-centric policy learning presented in the previous section is not a trivial task. However, recent progress in computer vision and especially in generalist vision foundation models, makes this possible now. In this section, we present a perception pipeline that can detect and track affordance frames. 

We note that we specifically do not claim the following approach to be superior to alternative methods. We do not present an in-depth analysis or comparison to alternatives, as this is beyond the scope of our paper. We found that the pipeline presented in the following was effective in our setup, but future work definitely can improve upon what we present here.
A complete visual overview of our perception pipeline is given in Fig.~\ref{fig:main}.

\textbf{Assumptions: } We assume that for each task, the human demonstrator provides the following input during the policy training stage to aid the perception system later during autonomous policy execution: (1) the name of the target object the robot has to interact with, e.g.\ ``\textit{teapot}''; and (2) three reference points that localise the affordance-frame of the task.


\textbf{Affordance Frame Detection: }
At the start of each subtask, given an input image of the robot’s workspace from an external camera (see Fig.~\ref{fig:robot}) and the user-provided name of the target object, we first use Grounding DINO~\cite{liu2023grounding} to detect a bounding box around the object, leveraging its open-vocabulary capabilities. The image is then cropped to this bounding box and passed to SAM~\cite{kirillov2023segment} to obtain a segmentation mask of the object. We additionally pass the cropped image to CSM Cube’s~\cite{csmcube} Image-to-3D model, which generates a textured mesh of the object. 
Both the segmentation mask and the textured mesh are then used to initialise Foundation Pose~\cite{wen2023foundationpose} for object pose estimation. 

However, this pose is extracted relative to the mesh centre rather than the specific affordance relevant to the task. To localise the affordance pose, we extract DINO-ViT features~\cite{amir2021deep} from the current image crop and match them against the DINO-ViT features of the three reference points provided by the human demonstrator during policy training. This matching process allows us to transform the mesh-centred pose detected by Foundation Pose to align with the affordance region. This process is illustrated in Fig.~\ref{fig:main} (top).

This entire process is performed once at the start of the task. For long-horizon tasks involving multiple objects and sequential subtasks, we initialise all affordances at the beginning and track them in parallel using multiple Foundation Pose models.

\textbf{Affordance Frame Tracking: }
Once initialised, we continuously track the pose of an object using Foundation Pose~\cite{wen2023foundationpose} and transform it to the affordance region at approximately 20~Hz on a desktop computer with an RTX4090 GPU. This tracked frame is used to compute the oriented affordance frame in which states and actions are represented. As we switch between subtasks for long-horizon tasks, we transition between the already-initialised affordance frames required for the respective subtasks.

\textbf{Efficacy of Perception Pipeline: }
In the experiments, we decoupled the effects of the perception pipeline (described in Appendix ~\ref{sec:perception}) on the policy performance and used fiducial markers on the objects to mark the pose of the various affordance frames. In this third set of experiments (Table \ref{tab:errors}), we show that our proposed perception pipeline is able to detect and track affordance frames without the use of fiducial markers, with only minimal decrease in task performance. 

\begin{figure*}[h]
    \centering
    \includegraphics[width=0.9\textwidth]{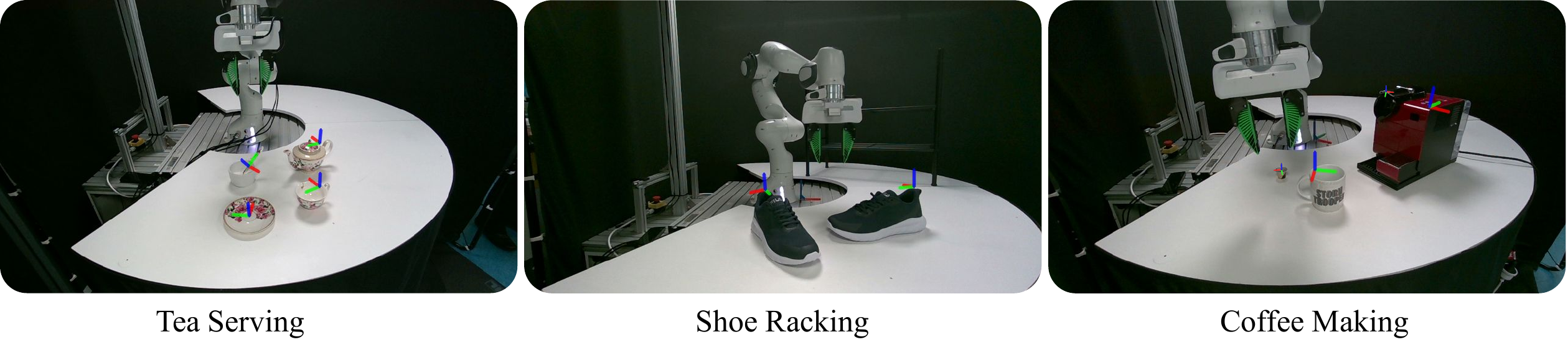}
    \caption{\small{\textbf{Demonstrating our system across three diverse real-world tasks.} The Tea Serving and Coffee Making tasks are very complex compared to many tasks typically encountered in the literature and require multiple sequential object interactions and a high level of interaction precision, e.g. when operating the coffee machine or pouring tea into the cup. Videos of the robot autonomously executing learned policies for all tasks are provided in the supplementary material.}} 
    \label{fig:tasks}
\end{figure*}

We ran 10 trials of the \emph{cup rearrange} task, which is a composition of the base tasks of rotating and grasping the tea cup and placing it on the saucer. When using the fiducial markers, 8 out of 10 trials were successful, which is consistent with the results reported in Table~\ref{tab:sub-policy-eval}, as expected. Removing the fiducial markers and using the proposed perception pipeline to detect and track the affordance frames resulted in 7 successful trials, indicating that only one additional failure case was introduced by the full perception pipeline. This failure case was a tracking error, where the vision system lost track of the objects pose due to occlusions from the robot, whereas the other two failure cases were a result of the policy failing to successfully grasp the cup or stagnation typically seen when the policy falls out of distribution. Further qualitative results in the accompanying videos show that we can train policies from 10 demonstrations and successfully execute them without any fiducial markers for the full tea-serving task, as well as the shoe-racking and coffee-making tasks, as illustrated in Fig.~\ref{fig:tasks}. 

\begin{table}[]
\centering
\resizebox{0.65\textwidth}{!}{%
\begin{tabular}{@{}lccccc@{}}
\toprule
& \textbf{Success Rate} & \multicolumn{3}{c}{\textbf{Type of Error}} \\ 
\midrule
& &
\shortstack{\textbf{Joint}\\\textbf{Limit Violation}} &
\shortstack{\textbf{Out of}\\\textbf{Distribution}} &
\shortstack{\textbf{Tracking}\\\textbf{Error}} \\
\midrule
\rowcolor[HTML]{EFEFEF} 
\textbf{Affordance Frames} & & & & \\
Oriented                   & \textbf{82\%} & 0.0\%  & 100\%  & 0.0\%  \\
Non-Oriented               & \textbf{46\%} & 36.4\% & 63.6\% & 0.0\%  \\
\midrule
\rowcolor[HTML]{EFEFEF} 
\textbf{Perception System} & & & & \\
Aruco Markers              & \textbf{80\%} & 0.0\%  & 100\%  & 0.0\%  \\
Foundation Pose \cite{wen2023foundationpose} & \textbf{70\%} & 0.0\%  & 66.7\% & 33.3\% \\ 
\bottomrule\\
\end{tabular}
}
\caption{\small{\textbf{Ablation Study.} Analysis of failure modes when comparing the two different affordance frames and perception systems.}}
\label{tab:errors}
\end{table}

\subsection{Assumptions and Limitations}
\label{sec:limitations}
There is no free lunch~\cite{wolpert1996lack}, and the reduction in required demonstrations while gaining spatial and intra-category invariance does not come for free. While our proposed approach significantly reduces the burden on the human demonstrator to provide a large number of task demonstrations, we make the following assumptions: 

(1) The objects involved in the tasks have clearly defined affordances, and a human demonstrator can identify the location of the relevant affordance frames. We find this to be a mild assumption for many objects involved in typical tasks in a domestic scenario, but we acknowledge that some tasks (e.g. laundry folding) will break this assumption. 

(2) The affordance frames appear at locations on the object that are distinct and informative enough for a perception pipeline to identify and track them, as well as transfer them across objects within the same category. We found this assumption to hold well for the evaluated real-world tasks, but objects without characteristic geometries or appearance will pose a challenge. 

(3) The human demonstrator can identify sub-tasks within the long-horizon task. This is a very mild assumption, and the partitioning of tasks could be automated based on detecting when the robot starts interacting with the next object, or even with the help of Large Language Models.

\textbf{Limitations: }
While our evaluation showed the proposed approach to be effective in learning long-horizon tasks, there are some noteworthy limitations that warrant further exploration. Most importantly, the proposed method depends on reliable object tracking to continually update the pose of the affordance frame during a manipulation task. While the presented perception pipeline from Sec.~\ref{sec:perception} worked well in the tested scenarios, it has limitations when tracking through occlusions, or dealing with non-rigid (e.g. articulated or deformable) objects.

Second, the pose-based abstraction of objects could limit the applicability to tasks not easily represented by object affordance frames alone, potentially requiring additional modalities like tactile sensing to capture more fine-grained object details. Despite these limitations, our contribution offers a promising pathway to more sample-efficient and generalisable imitation learning of complex long-horizon manipulation tasks.

Informally, our approach relieves the human demonstrator from the burden of collecting a large number of diverse demonstrations and reduces the pressure on the policy learning algorithm to extract task-relevant generalisation information from raw image-based state inputs. Instead, we shift part of the inherent difficulty of imitation learning to a dedicated perception system that can extract and track affordance frames. One might argue that this merely redistributes the difficulty rather than reducing it. However, we are confident that this shift is highly beneficial: acquiring large-scale training data for generalist vision models, such as those used in Sec.~\ref{sec:perception}, is significantly cheaper and more scalable than collecting extensive human demonstrations and robot interaction data, which remain expensive and labour-intensive. Thus, we expect the performance of specialised vision perception systems to continue to improve rapidly, becoming even more useful in the context of imitation learning soon. 


\subsection{Applicability to Mobile Manipulation}

By training our policy with respect to a relative frame attached to an object, the robot's action and state space remain consistent regardless of the position of the robot's base. This allows for the policy to continue operation while the base of the robot is in motion. We demonstrate this by running the same policy trained in the tabletop setting on a mobile manipulator robot and show how the end effector of the robot can maintain task performance regardless of the movement of the robot's base as illustrated by the discrepancy between the green and red robot base locations in Figure \ref{fig:mobile_base}. A video of this experiment is provided in the supplementary material.

\begin{figure*}[tbh]
    \centering
    \includegraphics[width=1.0\textwidth]{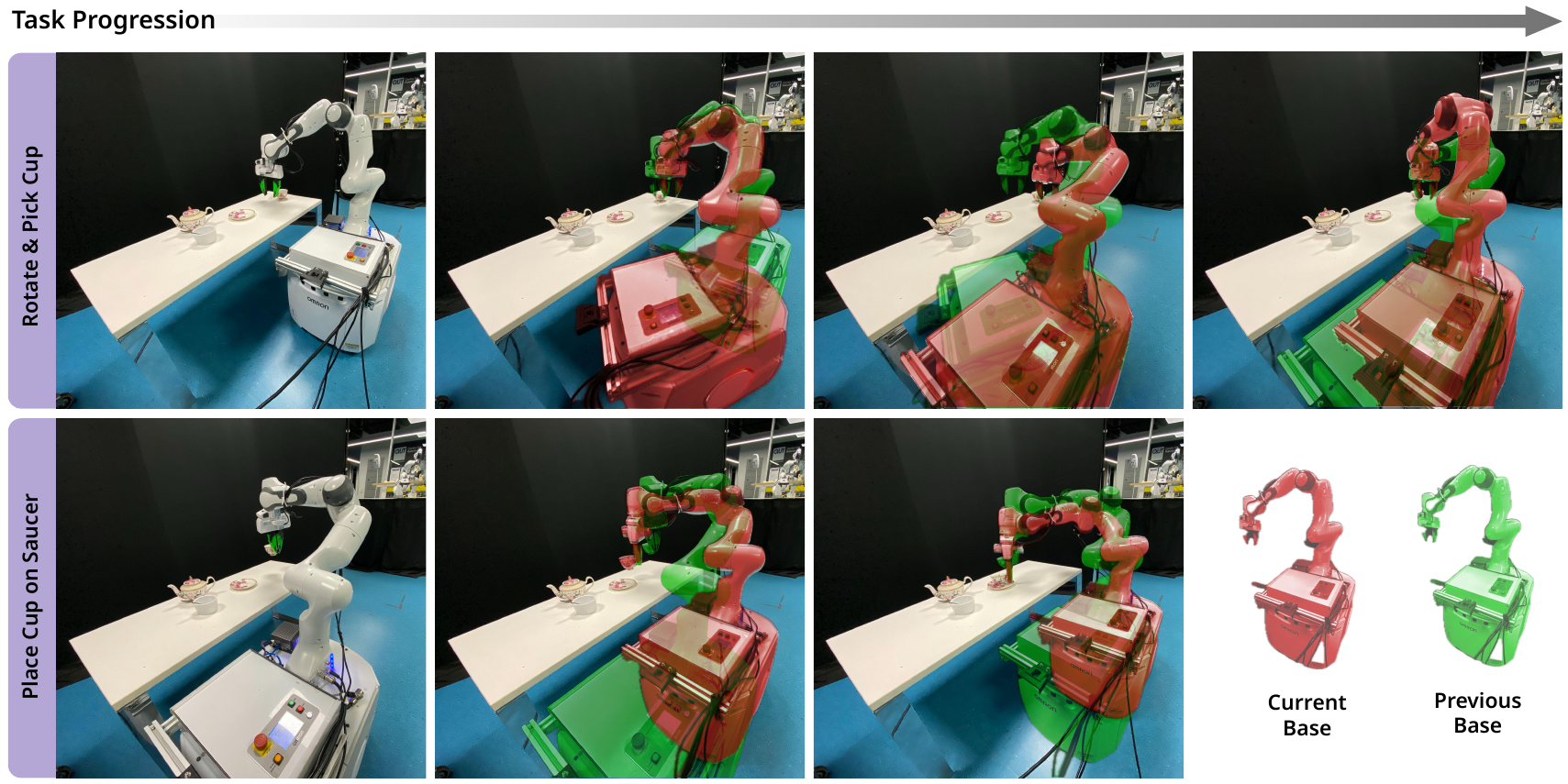}
    \caption{\textbf{Robustness to moving base.} We demonstrate our ability to maintain task performance regardless of the robot's moving base when operating with respect to an affordance-centric task frame.
    }
    \label{fig:mobile_base}
\end{figure*}

\subsection{Closed Loop Control}

Prior work have introduced the concept of local keypoints or regions on objects as compact representations for manipulation tasks \cite{simeonovdu2021ndf, manuelli2019kpam, dipalo2024kat}. These systems have traditionally been used to define start and end poses for simple pick-and-place operations, utilizing off-the-shelf inverse kinematics and motion planners to move objects from one location to another in an open-loop manner \cite{simeonovdu2021ndf,manuelli2019kpam}. Other methods have incorporated these representations within the context of imitation learning, primarily focusing on one-shot imitation learning \cite{gao2023k, gao2024bi, heppert2024ditto}. In these cases, the keypoint locations are used to define complex admittance controllers \cite{gao2023k} or prompt large language model (LLM) \cite{dipalo2024kat} to replicate a single trajectory, limiting their ability to react to changes or perturbations during policy execution. Our approach in contrast, leverages these representations in a behaviour cloning setting where we can learn closed-loop diffusion policies \cite{chi2023diffusionpolicy} that are robust to perturbations and allow us to move beyond simple pick-and-place tasks to imitating more complex closed-loop tasks, including non-prehensile manipulation, such as pushing objects as shown in Figure \ref{fig:closed_loop} below.

\begin{figure*}[tbh]
    \centering
    \includegraphics[width=1.0\textwidth]{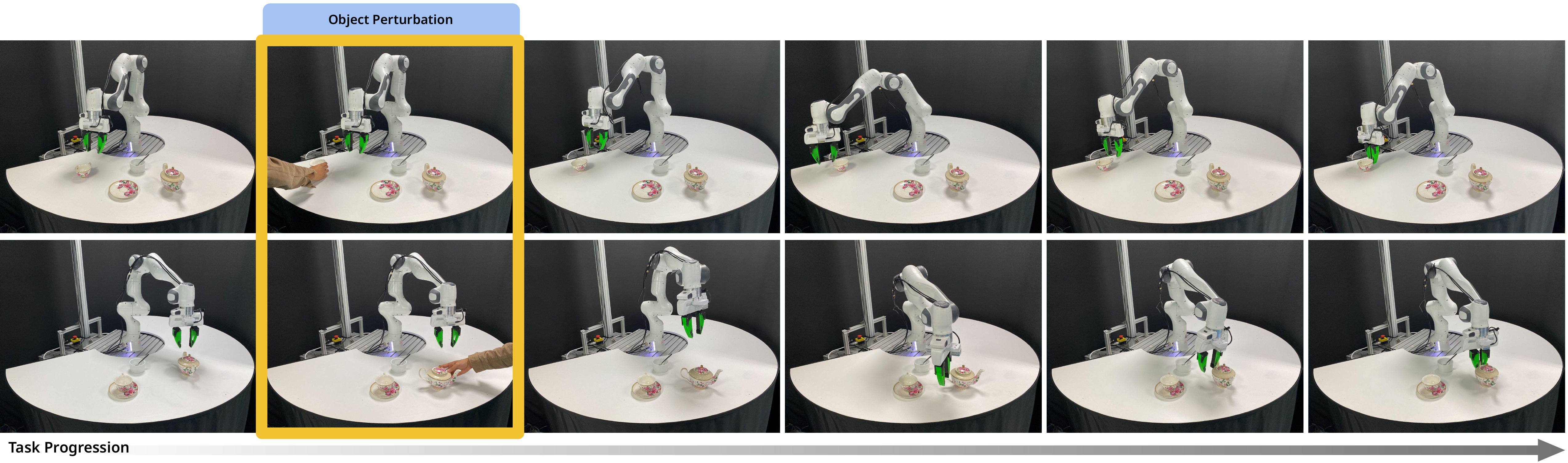}
    \caption{\textbf{Closed-loop control.} We demonstrate our ability to learn robust policies that react to object perturbations during execution beyond simple pick-and-place tasks.}
    \label{fig:closed_loop}
\end{figure*}


\subsection{Tasks}

\subsection{Experimental Setup -- Further Details}
\begin{figure}[t]
    \centering
    \includegraphics[width=1.0\linewidth]{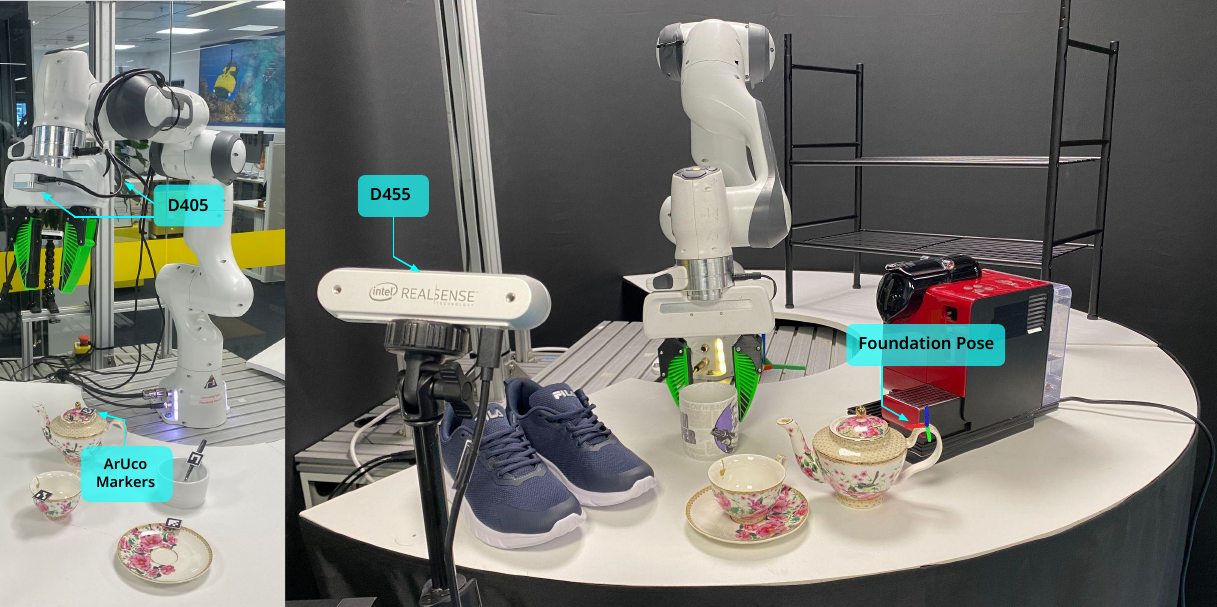} 
    \caption{\textbf{Experimental Setup.} \textit{Left:} Marker-based setup using 2 wrist-mounted D405 Intel Realsense cameras for top down detection. \textit{Right:} Marker-free setup using a front-facing D455 Intel Realsense camera running Foundation Pose.}
    \label{fig:robot}
\end{figure}

For all evaluations, we used a Franka Panda manipulator arm 
equipped with Intel Realsense cameras as shown in Figure~\ref{fig:robot}. We utilised Cartesian impedance control to control the robot. All demonstrations were collected using a GELLO teleoperation device~\cite{wu2023gello}. During data collection, all objects are equipped with an individual ArUco marker (Figure~\ref{fig:robot}) which we use for identifying the affordance-centric frames via measured rigid transforms from this marker, as well as for tracking these frames across the demonstration. We chose this method to obtain the affordance frame as it allowed us to decouple the performance of the perception system from the utility of the affordance frames for policy learning and composition, which was the main focus of this work. For all the video demonstrations in the supplementary, we switched to the marker-free setup as shown in Figure~\ref{fig:robot} (right). 

\subsection{Policy Composition -- Further Details}
\label{app:policy_comp}
Having trained each affordance-centric policy, we can compose them to solve long-horizon, multi-object tasks. We first define the order of sub-tasks and their associated affordance frames required to complete the full task. The robot then performs an initialisation scan of the environment to identify the initial locations of all objects and their local affordance frames in the scene. Once identified it runs the first policy corresponding to the first sub-task. As the policy is trained to output end-effector poses defined in the oriented affordance frame ${}^{\text{o-aff}}\vT_{\text{ee}}$, we transform these actions to the base frame of the robot $\vT_{\text{ee}}$ before executing them with a Cartesian impedance controller. If $a_\text{progress}$ generated by the policy increases beyond a predefined threshold $\phi$, indicating sub-task completion, we switch affordance frames and repeat the process with the next policy corresponding to the next sub-task.

\begin{figure*}[t]
    \centering
    \includegraphics[width=1.0\textwidth]{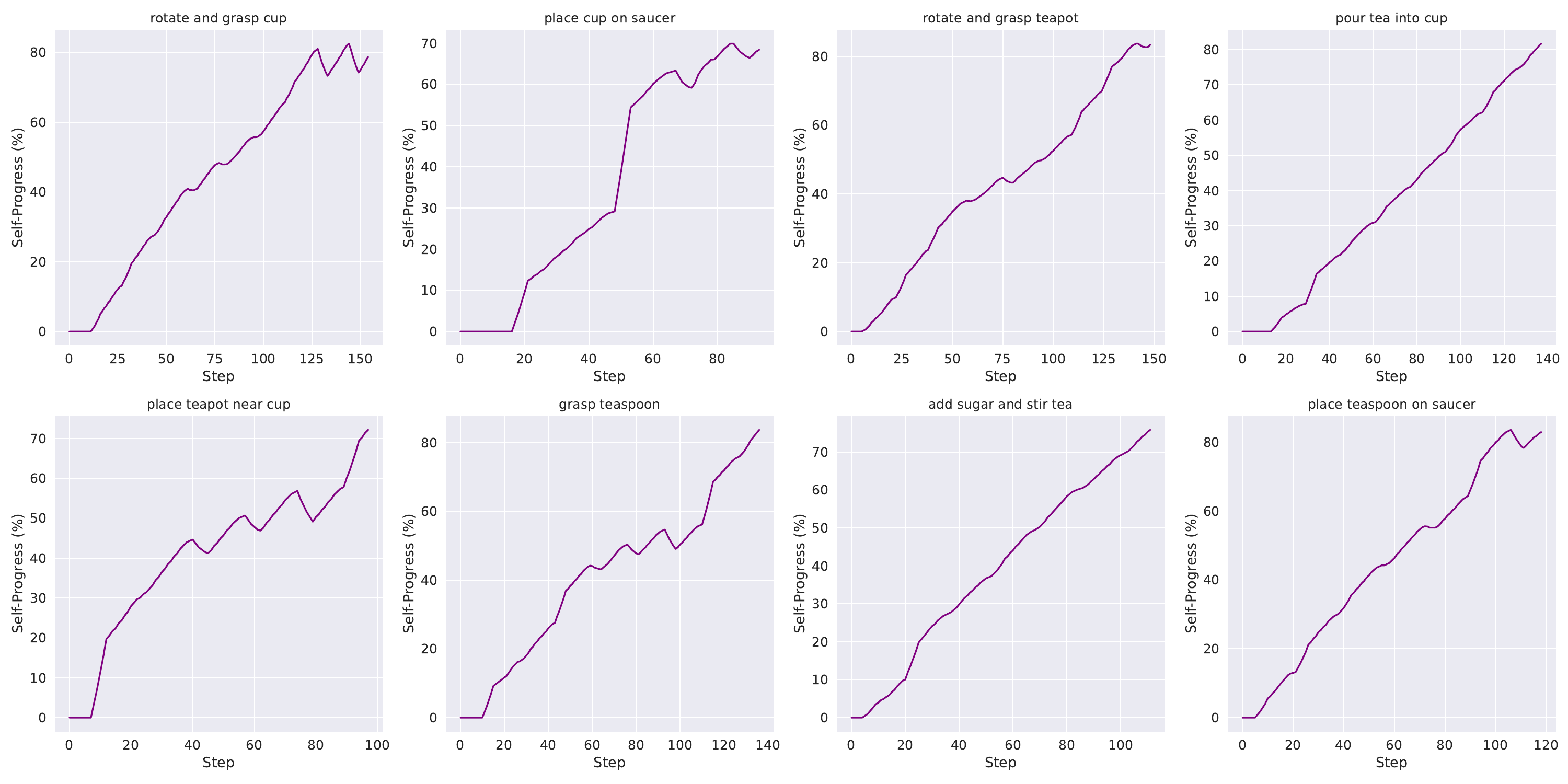}
    \caption{\small{\textbf{Self-Progress Predictions across the Tea Making Task.} Each sub-plot indicates the predicted self-progress for a different sub-task in the tea-making task.}}
    \label{fig:progress_abalate_all}
\end{figure*}

Figure \ref{fig:progress_abalate_all} illustrates the predicted progress while executing each base task in the tea-serving scenario. The system switches to the next policy when the predicted progress reaches a predefined threshold.

\begin{figure*}[t]
    \centering
    \includegraphics[width=1.0\textwidth]{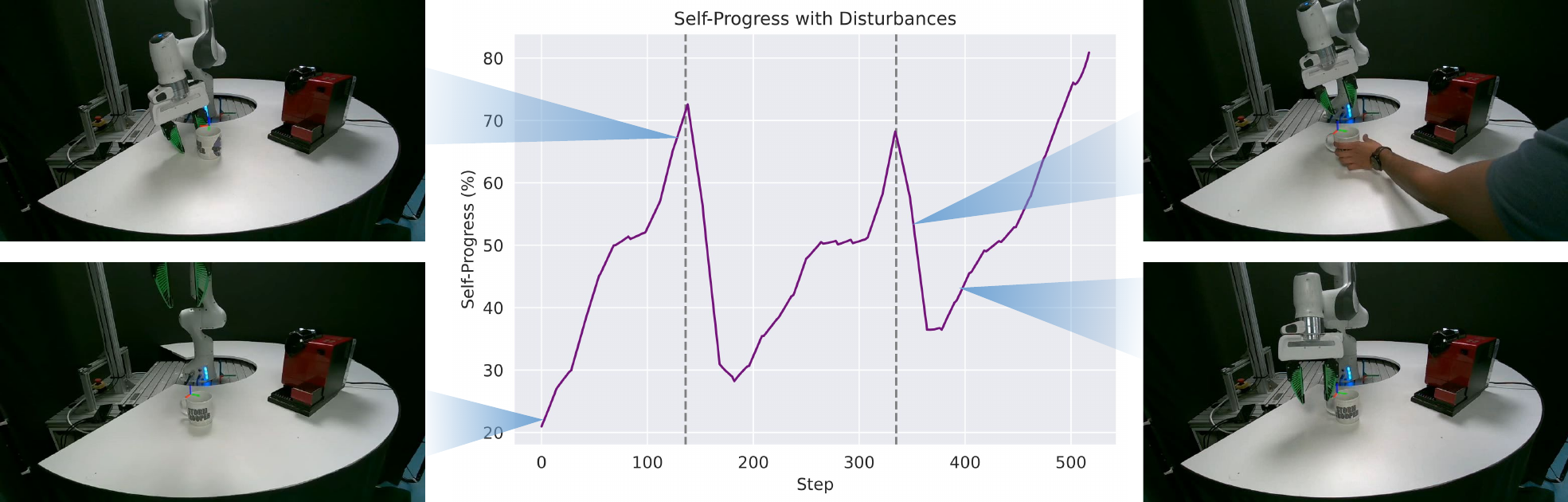}
    \caption{\small{\textbf{Self-Progress Behaviour with Disturbances.} Predicted progress over the course of a task exposed to two different disturbances and resets mid-execution. The task requires the robot to rotate a coffee mug and then grasp it. We indicate the start of the two disturbances where we reset this rotation by the grey dotted vertical lines in the plot. }}
    \label{fig:progress_disturb}
\end{figure*}

We further tested the responsiveness of this self-progress indicator to external disturbances for the task of rotating a cup and then grasping it. As illustrated in Fig.~\ref{fig:progress_disturb}, a human interfered during task execution by moving or rotating the cup. The self-progress prediction value immediately decreases as the task is partially reset and gradually increases as the robot proceeds with solving the task. 

\subsection{Diffusion Policy}

Throughout this work, we leverage diffusion policies~\cite{chi2023diffusionpolicy} as our central behaviour cloning algorithm. Diffusion policy models the conditional action distribution as a denoising diffusion probabilistic model (DDPM), allowing for better representation of the multi-modality in human-collected demonstrations. Specifically, diffusion policy uses DDPM to model the action sequence $p(\mathbf{A}_t \mid \mathbf{o}_t, \mathbf{x}_t)$, where $\mathbf{A}_t = \{\mathbf{a}_t, \dots, \mathbf{a}_{t+C}\}$ represents a chunk of next $C$ actions. The final action is output of the following denoising process:
\begin{equation}
    \mathbf{A}_t^{k-1} = \alpha (\mathbf{A}_t^{k} - \gamma \epsilon_\theta(\mathbf{o}_t, \mathbf{x}_t, \mathbf{A}_t^{k})) + \mathcal{N}(0, \sigma^2 \mathbf{I}),
\end{equation}
where $\mathbf{A}_t^{k}$ is the denoised action sequence at time $k$. Denoising starts from $\mathbf{A}_t^{K}$ sampled from Gaussian noise and is repeated till $k = 1$. In Equation (1), $(\alpha, \gamma, \sigma)$ are the parameters of the denoising process and $\epsilon_\theta$ is the score function trained using the MSE loss $\ell(\theta) = (\epsilon_k - \epsilon_\theta(\mathbf{o}_t, \mathbf{x}_t, \mathbf{A}_t^{k} + \epsilon_k))^2$. The noise at step $k$ of the diffusion process, $\epsilon_k$, is sampled from a Gaussian of appropriate variance.

The policy predicts a sequence of 16 actions, of which we execute the first 8. The diffusion network has 8.08 million parameters, we used a learning rate of $10^{-4}$.
The rest of the implementation is identical to the original implementation~\cite{chi2023diffusionpolicy}. 

\begin{wrapfigure}{l}{1.0\columnwidth} 
  \vspace{-10pt} 
  \begin{algorithm}[H]
    \caption{Calculation of $\vR_{\text{align}}$}
    \SetAlgoLined
    \KwIn{$\mathbf{p}_{\text{tool}}, \mathbf{p}_{\text{afford}}$}
    \KwOut{$\vR_{\text{align}}$}
    \SetKwFunction{FMain}{ComputeRotationMatrix}
    \SetKwProg{Fn}{Function}{:}{}
    \Fn{\FMain{$\mathbf{p}_{\text{tool}}, \mathbf{p}_{\text{afford}}$}}{
        \textbf{Define the Vectors:}\\
        $\mathbf{v}_{\text{funnel}} \gets [1, 0, 0]^T$\\
        $\mathbf{p}_{\text{tool}} \gets$ Position of the tool frame\\
        $\mathbf{p}_{\text{afford}} \gets$ Position of the affordance frame\\
        \textbf{Calculate the Direction Vector:}\\
        $\mathbf{d} \gets \mathbf{p}_{\text{tool}} - \mathbf{p}_{\text{afford}}$\\
        $\mathbf{d}_{\text{norm}} \gets \frac{\mathbf{d}}{\|\mathbf{d}\|}$\\
        \textbf{Find the Rotation Axis and Angle:}\\
        $\mathbf{r} \gets \mathbf{v}_{\text{funnel}} \times \mathbf{d}_{\text{norm}}$\\
        $\mathbf{r}_{\text{norm}} \gets \frac{\mathbf{r}}{\|\mathbf{r}\|}$\\
        $\cos(\theta) \gets \mathbf{v}_{\text{funnel}} \cdot \mathbf{d}_{\text{norm}}$\\
        $\sin(\theta) \gets \|\mathbf{r}\|$\\
        \textbf{Construct the Rotation Matrix:}\\
        $\mathbf{K} \gets \begin{bmatrix}
        0 & -r_z & r_y \\
        r_z & 0 & -r_x \\
        -r_y & r_x & 0
        \end{bmatrix}$\\
        $\vR_{\text{align}} \gets I + \sin(\theta) \mathbf{K} + (1 - \cos(\theta)) \mathbf{K}^2$\\
        \KwRet{$\vR_{\text{align}}$}}
  \end{algorithm}
  \label{algorithm1}
  \vspace{-10pt} 
\end{wrapfigure}